\documentclass[twoside]{article}

\usepackage{aistats_arxiv}
\usepackage[utf8]{inputenc} 
\usepackage[T1]{fontenc}    
\usepackage{hyperref}       
\usepackage{url}            
\usepackage{booktabs}       
\usepackage{amsfonts}       
\usepackage{nicefrac}       
\usepackage{microtype}      
\usepackage{amsmath}
\usepackage{amssymb}
\usepackage{subfiles}
\usepackage{graphicx}
\usepackage{subcaption}
\usepackage{multirow}
\usepackage{tikz}
\usepackage{centernot}
\usepackage{framed}
\usepackage{authblk}

\usetikzlibrary{shapes,backgrounds}

\usepackage{algorithm}
\usepackage{algorithmicx}

\usepackage{amsthm}       

\newtheorem{corollary}{Corollary}
\newtheorem{proposition}{Proposition}

\theoremstyle{definition}

\newcommand\independent{\protect\mathpalette{\protect\independenT}{\perp}}
\def\independenT#1#2{\mathrel{\rlap{$#1#2$}\mkern2mu{#1#2}}}

\begin{document}

%


\title{Partial Identifiability in Discrete Data With Measurement Error}



\author[1]{Noam Finkelstein\thanks{Equal Contribution}}
\author[1]{Roy Adams$^{*}$}
\author[1,2,4]{Suchi Saria}
\author[1,3]{Ilya Shpitser}
\affil[1]{\footnotesize Department of Computer Science, Johns Hopkins University}
\affil[2]{\footnotesize Department of Applied Mathematics and Statistics, Johns Hopkins University}
\affil[3]{\footnotesize Department of Biostatistics, Johns Hopkins University}
\affil[4]{\footnotesize Bayesian Health}
\date{}

\maketitle


\begin{abstract} 
	
  When data contains measurement errors, it is necessary to make assumptions
  relating the observed, erroneous data to the unobserved true phenomena of
  interest. These assumptions should be justifiable on substantive grounds, but
  are often motivated by mathematical convenience, for the sake of exactly
  identifying the target of inference. We adopt the view that it is preferable
  to present bounds under justifiable assumptions than to pursue exact
  identification under dubious ones. To that end, we demonstrate how a broad
  class of modeling assumptions involving discrete variables, including common measurement error and
  conditional independence assumptions, can be expressed as linear constraints
  on the parameters of the model. We then use linear programming techniques to
  produce sharp bounds for factual and counterfactual distributions under
  measurement error in such models. We additionally propose a procedure for
  obtaining outer bounds on non-linear models. Our method yields sharp bounds in
  a number of important settings -- such as the instrumental variable scenario
  with measurement error -- for which no bounds were previously known.
\end{abstract}

%

\section{Introduction}

Measurement error is a ubiquitous problem in fields ranging from epidemiology
and public health \cite{rothman2008modern,adams2019learning} to economics
\cite{horowitz1995contaminated,imai2010treatment,molinari2008misclassified} to
ecology \cite{solymos2012conditional,shankar2019three}. If unaccounted for
during analysis, measurement error can lead biased results. Accounting for
measurement error bias requires making assumptions about how errors occur so as
to link the available data to the underlying true values. As such, it is
important to the validity of the analysis and the resulting conclusions that
these assumptions be substantively justifiable.

In practice, however, analysts often make implausibly strong assumptions for
the sake of \emph{exactly identifying} their target of inference. In such
cases, we advocate instead for the use of weaker, credible assumptions to
identify \emph{bounds} on the target of inference, referred to as \emph{partial
identification}. Previous work on partial identifiability under measurement
error has largely focused on deriving analytic bounds for
specific combinations of settings and assumptions. As a result, there remain
many important and common settings for which no known bounds exist.

In this work, we adopt a latent variable formulation of the measurement error
problem. That is, we are interested in estimating a \textbf{target parameter}
that involves the distribution of a discrete unobserved variable $X$ using
observations of a discrete observed proxy variable $Y$. We show how sharp
partial identification bounds for the target parameter can be computed
numerically and without extensive derivations for a class of models, including
several models for which no analytical bounds are currently known. Our
approach, which is similar in spirit to that of \cite{balke93ivbounds}, is to
encode the target parameter and constraints imposed by the model as a
linear program which can be maximized and minimized to produce sharp bounds. We
show that this approach can be used to compute bounds for factual and
counterfactual parameters of a discrete mismeasured variable, including its
marginal distribution and its moments, and the average treatment effect (ATE)
of an intervention on the mismeasured variable.

The primary contribution of this paper is a collection of modeling assumptions
that can be encoded as linear constraints on a target parameter and, thus, are
amenable to the linear programming approach. These include several common
\textbf{measurement error} constraints (Section \ref{sec:base}),
\textbf{graphical} constraints encoded by hidden variable Bayesian networks
(Section \ref{sec:iv}), and \textbf{causal} constraints relating potential
outcomes under different interventions (Section \ref{sec:causal-bounds}). Our
main result, presented in Section \ref{sec:iv}, draws on results from
\cite{fine} and \cite{evans} to show how an extended class of instrumental
variable-type models produce linear constraints on a target parameter and
present a simple procedure for obtaining outer bounds for graphs outside this
class. 


Our approach allows a practitioner to mix and match any combination of the
modeling assumptions described in this paper with little effort. This
flexibility means that it is trivial to compute bounds under new models and to
test the sensitivity of the resulting bounds to different combinations of
assumptions. Throughout, we demonstrate the utility of this approach on
synthetic examples representing important use cases. Our goal in these examples
is to demonstrate how to apply the approach to commonly occurring models so that
practitioners may adapt them to their specific data scenario. These cases
include:

\begin{enumerate}
\item Bounding the mean of a variable using a single noisy proxy
\item Bounding the mean of a variable using two noisy proxies with an exclusion
  restriction
\item Bounding the ATE in a randomized trial with measurement error on the
  outcome
\item Bounding the ATE in an instrumental variable model with measurement error
  on the outcome
\end{enumerate}

These cases represent common scenarios occurring in economics, public health,
political science, and other scientific settings. Among these, cases 2, 3, and 4
all represent settings for which no sharp symbolic bounds are currently known
and previous work suggests that deriving such symbolic bounds automatically for
these settings is infeasible \cite{bonet2001instrumentality}. We begin by
introducing our approach in the context of the simplest possible measurement
error model (Section \ref{sec:base}). Next, we describe a class a class of graphical models that yield linear constraints



%
	
\section{The basic measurement error problem and the LP formulation}
\label{sec:base}

Suppose that we are interested in the distribution of a discrete random
variable $X \in \mathcal{X}$, but instead of observing $X$ directly, we can
only observe a discrete noisy proxy, denoted $Y \in \mathcal{Y}$. Clearly,
without any assumptions about the proxy distribution, $P(Y|X)$, we cannot say
anything about the distribution of $X$. On the opposite extreme, if $P(Y|X)$ is
known and invertible, then $P(X)$ is fully identifiable from observations of
$Y$ \cite{rothman2008modern}.


Our interest is in between these two extremes, where we assume certain
properties of $P(Y|X)$, but not the whole distribution. Our goal is to bound
functions of $P$, referred to as \emph{parameters of interest}, under these
assumptions. Our approach is to translate the modeling assumptions into a set
of constraints on $P(X,Y)$ and bound the parameter of interest by finding its
maximum and minimum subject to these constraints. Formally, let $\Delta^{d}$ be
the $d$-dimensional simplex, let $\eta: \Delta^{|\mathcal{X}|\times |\mathcal{Y}|} \to \mathbb{R}$
be the parameter of interest, and let $\mathcal{M} \subseteq
\Delta^{|\mathcal{X}| \times |\mathcal{Y}|}$ be the set of distributions
allowed under the modeling assumptions. Then, assuming that $\mathcal{M}$
contains the true distribution $P_0$, we can bound $\eta(P_0)$ as
\begin{align}
	\min_{P\in\mathcal{M}} \eta(P) \leq \eta(P_0) \leq \max_{P\in\mathcal{M}} \eta(P)
\end{align}
We will show how to construct $\mathcal{M}$ such that this optimization problem
is tractable and the bounds produced are sharp.

%
For notational simplicity, let $\phi_{xy} := P(X = x, Y = y)$ be the joint
distribution of $X$ and $Y$. By construction, $\phi$ must satisfy the the
\textbf{probability constraints} $\sum_{xy} \phi_{xy} = 1$, and for all $x, y$,
$\phi_{xy} \ge 0$. In addition, $\phi$ must match the marginal for the observed
proxy $Y$, so that for all $y$, $\sum_{x} \phi_{xy} = P(Y=y)$. These are called
the \textbf{observed data constraints}.

As mentioned above, we must make assumptions about the relationship of the
unobserved variable $X$ and the observed proxy $Y$. We seek to avoid strong,
parametric assumptions, in favor of weaker assumptions that may be justified by
expert knowledge or domain research. Below, we provide examples of several such
commonly made assumptions. We call constraints on $\phi$ relating the proxies
to the unobserved variables the \textbf{measurement error constraints}
\footnote{ Each of these constraints can easily be soften by adding a slack
parameter which can, in turn, be varied in a sensitivity analysis. }.

\begin{enumerate}
  \item[(A0)] Bounded error proportion:
    $\sum_{x \neq y} \phi_{xy} \leq \epsilon$
  \item[(A1)] Unidirectional errors:
    $\sum_{y < x} \phi_{xy} = 0$
  \item[(A2)] Symmetric error probabilities:\\
    $~~~~~\phi_{xy} = \phi_{xy}\,\,\,\forall\,\,|x - y| = |x - y'|$
  \item[(A3)] Error probabilities decrease by distance:\\
    $~~~~~\phi_{xy} \geq \phi_{xy'}\,\,\,\forall\,\,|x - y| > |x - y'|$
\end{enumerate}

Assumption (A0) may be reasonable when there is sufficient previous literature to specify a range of plausible error rates, but not the exact proxy distribution (e.g., see discussion of sensitivity analysis in \cite{rothman2008modern}). Assumption (A1) may be used to represent positive label only data, which is common in areas such as ecology \cite{solymos2012conditional} and public health \cite{adams2019learning}. Assumption (A2) represents a generalization of the zero-mean measurement error assumption -- commonly made in settings where the errors are due to imprecision of the instrument -- to discrete variables. Finally, Assumption (A3) may be reasonable in settings where the values of $X$ have a natural ordering and small errors are more likely than large ones.

Note that the target distribution $P(X=x) = \sum_{y} \phi_{xy}$ and all
constraints described thus far can be written as linear functions of $\phi$. If
our parameter of interest is also a linear function of $\phi$, then the assumed
constraints, in combination with the linear objective corresponding to our
parameter of interest, form a linear program (LP) and bounds can be computed by
both maximizing and minimizing the parameter of interest using one of a number
of LP solvers. Parameters that are linear in $\phi$ include $P(X)$ and all of
its moments. By the intermediate value theorem, the objective can take any
value between its constrained maximum and minimum values. It follows that for
sets of assumptions that can be linearly expressed, the bounds obtained through
this linear program are sharp, i.e. they exhaust the information in the
observed data under the stated assumptions.

\begin{figure}[t]
	\begin{subfigure}[b]{0.49\textwidth}
			\includegraphics[width=\textwidth]{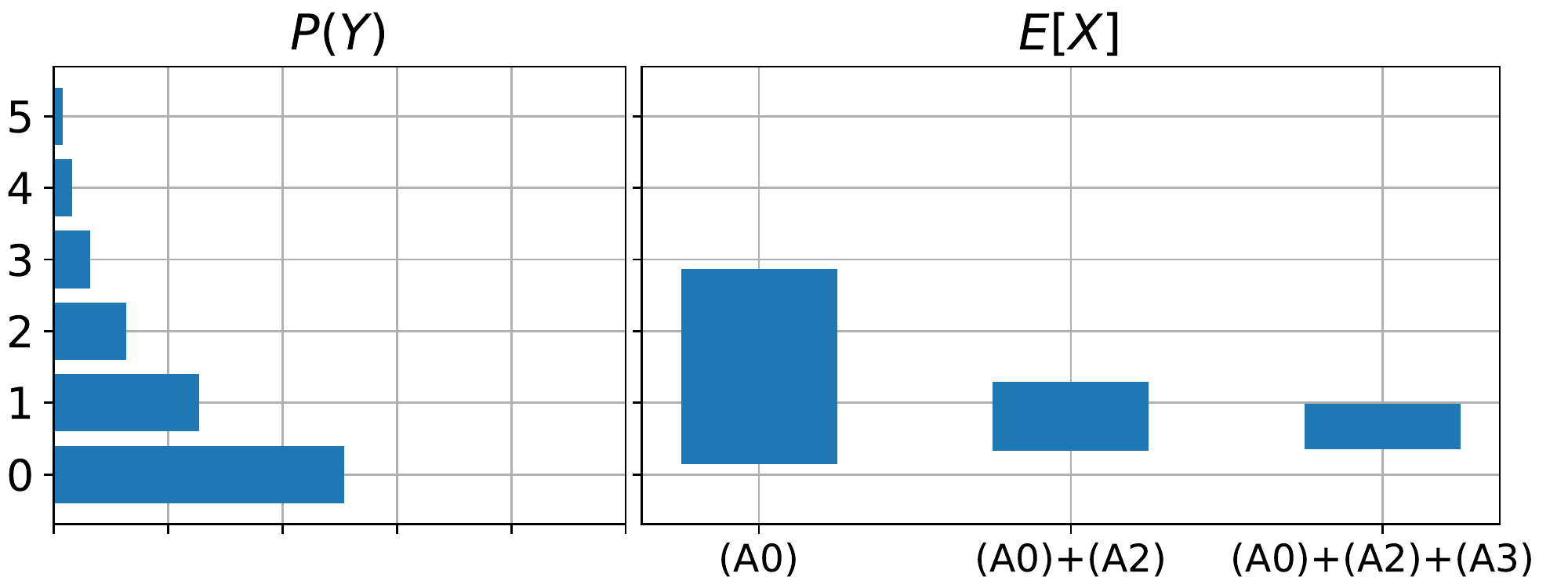}
    \end{subfigure}
    \begin{subfigure}[b]{0.49\textwidth}
			\includegraphics[width=\textwidth]{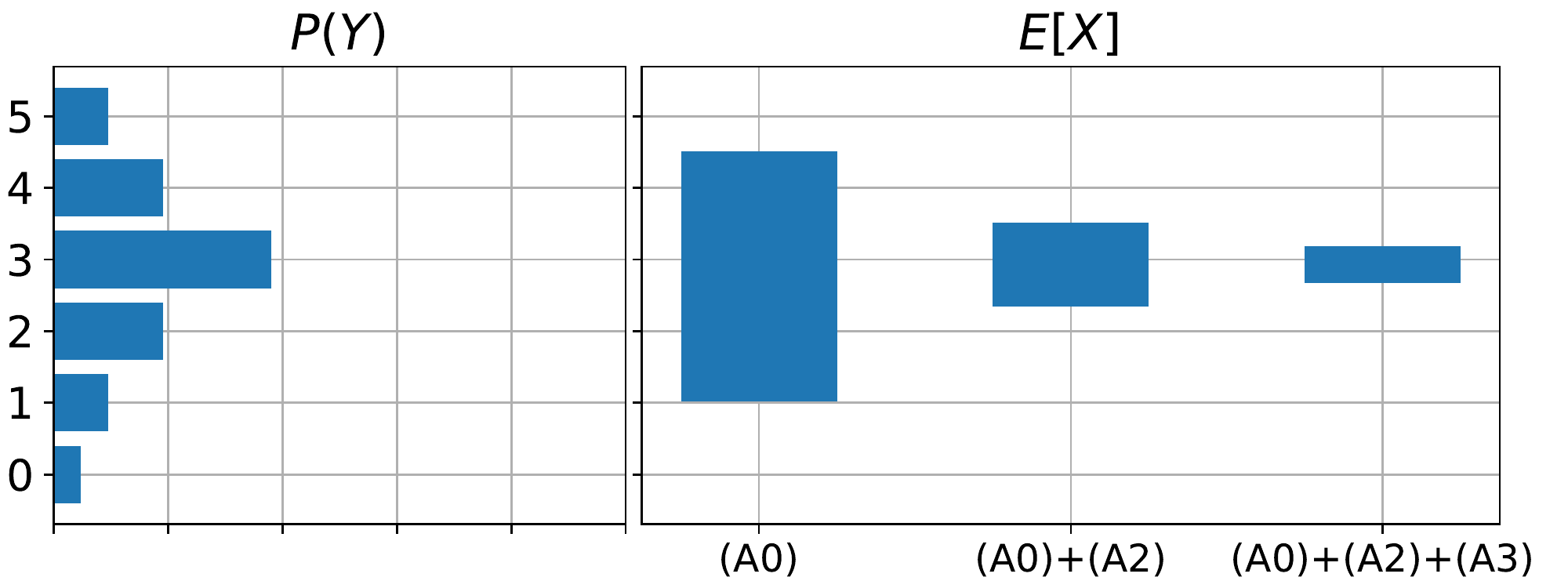}
	\end{subfigure}\hspace{1cm}
  \caption{
    Bounds on $E[X]$ under different measurement error assumptions and observed
    proxy distributions, $P(Y)$.
  }
  \label{fig:mrs_bounds}
\end{figure}

\subsection*{Application: Bounding $E[X]$ using a single proxy}

To demonstrate this approach, suppose we are interested in bounding the
expected value of a variable $X \in \{0,..,5\}$ from observations of $Y \in
\{0,...,5\}$. We will consider bounds under (A2), (A3), and a weaker version of
(A0) wherein there is a known upper bound, $\epsilon = 0.01$, on the proportion
of observations that are mistaken by more than $2$ points (errors of $2$ points
or less are unconstrained by this assumption). The resulting LP is shown below.
\begin{align}
\label{eq:base}
\text{objective:~~~~~~~} \sum_{x,y} &x \phi_{xy}\\\nonumber
\text{constraints:~~} \sum_x \phi_{xy} &= P(Y = y) \\\nonumber
 \phi_{xy} & \ge 0 \\\nonumber
 \sum_{|x - y| > 2} \phi_{xy} &\le \epsilon \\\nonumber
 \phi_{xy} &\geq \phi_{xy'} & &
                 \forall |x - y | < |x - y'| \\\nonumber
 \phi_{xy} &= \phi_{xy'} & &
                  \forall |x - y| = |x - y'|
\end{align}
For two example proxy distributions, $P(Y)$, Figure \ref{fig:mrs_bounds} shows
the resulting bounds on $E[X]$ under different combinations of (A0), (A2), and
(A3). Bounds under each combination of assumptions were computed by simply
solving a slightly different version of the LP in Equation \ref{eq:base},
highlighting the ease with which we can perform sensitivity analysis without
rederiving bounds under each new model. \hfill\qedsymbol

Were multiple proxies $\mathbf{Y} \equiv \{Y_1,...,Y_K\}$ observed with no
assumptions made about the relationship between them, each proxy would be
subjected to its own observed data constraints and potentially its own
measurement error constraints, depending on what knowledge is available about
the error process. The objective, and each of the other constraints, would then
simply be expressed on the margins of the full distribution $P({X, \bf Y})$,
which maintains linearity.

\subsection{Extending the LP approach to other parameters and models}

In the remainder of this paper, we show how the LP approach described in this
section can be extended to bound other parameters of interest and incorporate
other modeling assumptions. In the Section \ref{sec:iv}, we show how to
incorporate conditional independence assumptions, encoded in a graphical model,
relating $X$ to other observed and latent variables beyond $Y$. We define a
class of latent variable Bayesian networks which produce linear constraints and
show how to relax the constraints imposed by models not in this class to
produce valid outer bounds. In Section \ref{sec:causal-bounds}, we consider
bounds on parameters of the distribution of $X$ under an intervention $A$. In
this setting, our interest is in the potential outcome variable $X(a)$, defined
as the value $X$ would have taken had we intervened to set $A=a$. Our goal,
then, is to bound parameters of the distribution $P(X(a))$, including the
average treatment effect of $A$ on $X$. We show how to bound such parameters
and incorporate additional assumptions relating potential outcomes under
different interventions.


%

\section{Graphical constraints}
\label{sec:iv}

In the previous section we relied on domain knowledge about the joint
distribution of $X$ and its proxy $Y$. In this section, we describe how
assumptions encoded in a graphical model can be used to further constrain our
target parameter. In particular, we describe a class of graphs that result in
linear constraints on the target parameter and, thus, is amenable to the linear
programming approach introduced in the previous section. This class includes the common instrumental variable (IV) model, shown in Figure \ref{fig:iv} (a), as well as the various extensions of this model shown in Figure \ref{fig:gen_iv}.

Suppose that we assume a latent variable Bayesian network $\mathcal{G} =
(\mathbf{V},\mathbf{E})$, where $\mathbf V$ and $\mathbf E$ represent the
vertices and edges of the network, respectively. For a variable $V \in
\mathbf{V}$, let $Pa_\mathcal{G}(V)$ be the parents of $V$ in $\mathcal{G}$
and $Ch_\mathcal{G}(V)$ be the children of $V$ in $\mathcal{G}$. Additionally,
we refer to the set of observed and unobserved variables with known cardinality
as \textbf{endogenous variables}, denoted by $\mathbf{O}$, and the set of
unobserved variables with unknown cardinality (typically latent confounders) as
\textbf{exogenous variables}, denoted by $\mathbf{U}$. For example, consider
the Bayesian network in Figure \ref{fig:iv} (a) in which $A$, $X$, and $Y$ are
endogenous and $U$ is exogenous. In this model $A$ is commonly referred to as
an \emph{instrumental variable} (IV), and the model is referred to as the IV
model. The independencies encoded in this graph, namely $A \perp U$ and $A
\perp Y \mid A,U$, place constraints on the joint distribution $P(A,U,X,Y)$
which, in turn, places constraints on the target parameter. We refer to the
target parameter given by a graphical model as \textbf{graphical constraints}. As before, our goal is to maximize and minimize the target parameter $\psi$ subject to these constraints. 

In general, the independence constraints imposed by a Bayesian network on the
joint distribution of variables in the graph are non-linear. For example, the
simple Markov chain model shown in Figure \ref{fig:iv} (b) yields quadratic
constraints on $P(A,X,Y)$ (see Proposition \ref{prop:binary_bounds} for more
details). This makes optimizing over the constraint set difficult as, in
general, quadratic programming is NP-hard \cite{pardalos1991quadratic}.
Further, the complete set of latent variable graphical models that impose
linear constraints on $P$ is unknown; however, in the remainder of this
section, we describe a class of graphs, including commonly used graphs such as
the IV model, that do yield linear constraints. We will proceed by first illustrating how the basic IV model produces linear constraints on $P$ and then generalizing this result to a class of graphs using results from \cite{fine} and \cite{evans}.

\subsection{Constructing linear constraints from the IV model}
To construct this class, we start by considering the IV model shown in Figure
\ref{fig:iv} (a) which is known to place linear constraints on $P$
\cite{balke93ivbounds,bonet2001instrumentality}. In order to arrive at linear
constraints, we will not optimize directly over the joint distribution $\phi =
P(A,U,X,Y)$ as we did in Section \ref{sec:base}. Instead, we will optimize over
an equivalent potential outcome distribution. Recall that a potential outcome
variable $X(a)$ represents the value $X$ would have taken had we intervened to
set $A = a$. Then, let $\tilde{X} = (X(a))_{a \in \mathcal{A}}$ and $\tilde{Y}
= (Y(x))_{x \in \mathcal{X}}$ be the vectors of potential outcome variables for
$X$ and $Y$ given their endogenous parents. Finally, let $\psi$ be the joint
distribution over $\tilde{X}$, and $\tilde{Y}$ such that
$\psi_{\tilde{x},\tilde{y}} = P(\tilde{X}=\tilde{x},\tilde{Y}=\tilde{y})$.

Under the consistency assumption, $\psi$ is connected to the distribution over
endogenous variables by the linear map $P(x, y \mid a) = P(X(a) = x, Y(x) =
y)$, where the last term is obtained by marginalizing all other variables in
$\tilde{X}$ and $\tilde{Y}$ out of the distribution $\psi$. For an explicit
example of this marginalization, see Section \ref{app:iv_lp} in the
Supplement. As observed in \cite{bonet2001instrumentality} and
\cite{balke93ivbounds}, all independencies in IV graph are now given by $A
\perp \tilde{X}, \tilde{Y}$ which can be written as
\begin{align}
\psi_{a,\tilde{x},\tilde{y}} = P(A=a)\sum_{a'} \psi_{a',\tilde{x},\tilde{y}}\,\,\,\forall\,a,\tilde{x},\tilde{y},
\end{align}
which is linear in $\psi$ since $P(A)$ is identified from the data. As
$\mathcal{G}$ imposes no other constraints on $\psi$, and marginalization is a
linear operation, all constraints on the conditional distribution $P(x, y \mid
a)$ are similarly linear.

\subsection{Graphs involving multiple instruments}
This linearity result can be generalized to more complex graphs
involving multiple instruments using the following proposition adapted
from \cite{fine}:

\begin{proposition}[Fine's Theorem] \label{prop:general_iv}
	Let $\mathcal{G} = (\mathcal{V},\mathcal{E})$ be a latent variable Bayesian
  network. Suppose there exists an exogenous latent variable $\Lambda \in
  \mathcal{V}$ such that (1) all descendants of $\Lambda$ are children of
  $\Lambda$ and (2) all non-descendants of $\Lambda$ are observed, have exactly
  one child, and that child is in $Ch_\mathcal{G}(\Lambda)$. Let the children of
  $\Lambda$ be denoted by $\bf X$, and the non-descendants of $\Lambda$ by $\bf
  A$. Then all constraints imposed by $\mathcal G$ on $P({\bf X} \mid {\bf A})$
  are linear.
\end{proposition}

In such a graph, all variables in $\bf X$ are mutually confounded by $\Lambda$
and we refer to the variables in $\bf A$ as \textbf{instruments}. Note that
this class of graphs trivially includes the basic IV model (Figure \ref{fig:iv}
(a)) but also extends the basic IV model in two important ways. First, we can
now include multiple instruments as shown in Figure \ref{fig:gen_iv} (a) (e.g.,
see \cite{angrist1991does,poderini2020exclusivity}). Second, we can include
instruments for both $X$ and its proxy $Y$, which may occur when some aspect of
the measurement process is randomized, such as the order of responses in a
survey or the gender of an in-person surveyor as shown in Figure
\ref{fig:gen_iv} (b) (e.g., see \cite{becker1995effect,catania1996effects}).

%

To derive the set of linear constraints imposed by such a graph, we can
generalize the procedure described for the IV model as follows. Let $K$ be the
number of variables in ${\bf X} = Ch(\Lambda)$. Then, for each variable $X \in
\mathbf{X} = (X_i)_{i=1}^{K}$, let $\tilde{X}$ be the potential outcomes of $X$
under each joint setting of its endogenous parents and let $\mathbf{\tilde{X}}
= (\tilde{X}_i)_{i=1}^{K}$ be the set of all such potential outcomes.
As before, let $\psi_{\bf a,\bf \tilde x} = P(\bf A = \bf a, \bf \tilde X = \bf \tilde x)$ denote the joint distribution over the instruments and
potential outcomes. Because $P(\mathbf{A})$ is assumed to be known, the only relevant independency imposed by the graph is given by ${\bf A} \perp {\bf \tilde X}$, which can be written as
\begin{align}
	\psi_{\bf a,\bf \tilde x} = P(\bf A= \bf a)\sum_{\bf a'} \psi_{\bf a',\bf \tilde x}\,\,\,\forall\,\bf a, \bf \tilde x,
\end{align}
which is linear in $\psi$. Finally, $\psi$ can be linearly mapped to the distribution over endogenous
variables $\mathbf{O} = \{\mathbf{A},\mathbf{X}\}$ as $P(\mathbf{O} = {\bf o}) = P({\bf A} = {\bf a}, \tilde{X}_1({\bf
o}_{Pa(X_1)}) = x_1,...,\tilde{X}_K({\bf o}_{Pa(X_K)}) = x_K)$
where ${\bf o}_{Pa(X_i)}$ are the values of
$X_i$'s parents in $\bf o$. The right hand side is obtained through
the (linear) marginalization of all other potential outcomes out of $\psi$ and thus the constraints imposed by the model on the distribution of endogenous variables are linear. This class of graphs already contains several useful models, but we will next expand this class further to include graphs with non-randomized instruments.

\subsection{Graphs with non-randomized instruments}

\begin{figure}[t]
  \centering {\small
    \begin{tikzpicture}[>=stealth, node distance=1.0cm]
  \begin{scope}[xshift=0.5cm]
    \tikzstyle{vertex} = [
      draw, thick, ellipse, minimum size=4.0mm, inner sep=1pt
    ]

    \tikzstyle{edge} = [
      ->, blue, very thick
    ]

    \node[vertex, circle] (a) {$A$};
    \node[vertex, circle, gray] (x) [right of=a] {$X$};
    \node[vertex, circle] (y) [right of=x] {$Y$};
    \node[vertex, circle, red, opacity=.7] (u) [above of=x, xshift=0.5cm, yshift=-0.5cm] {$U$};
    \draw[edge] (a) to (x);
    \draw[edge] (x) to (y);
    \draw[edge, red, opacity=.7] (u) to (x);
    \draw[edge, red, opacity=.7] (u) to (y);

    \node[below of=x, yshift=0.3cm] (l) {$(a)$};
  \end{scope} 

  \begin{scope}[xshift=4.45cm]
    \tikzstyle{vertex} = [
      draw, thick, ellipse, minimum size=4.0mm, inner sep=1pt
    ]

    \tikzstyle{edge} = [
      ->, blue, very thick
    ]

    \node[vertex, circle] (a) {$A$};
    \node[vertex, circle, gray] (x) [right of=a] {$X$};
    \node[vertex, circle] (y) [right of=x] {$Y$};

    \draw[edge] (a) to (x);
    \draw[edge] (x) to (y);

    \node[below of=x, yshift=0.3cm] (l) {$(b)$};
  \end{scope} 
 
  \begin{scope}[yshift=-2.3cm]
    \tikzstyle{vertex} = [
      draw, thick, ellipse, minimum size=4.0mm, inner sep=1pt
    ]

    \tikzstyle{edge} = [
      ->, blue, very thick
    ]

    \node[vertex, circle] (z) {$Z$};
    \node[vertex, circle] (a) [right of=z] {$A$};
    \node[vertex, circle, gray] (x) [right of=a] {$X$};
    \node[vertex, circle] (y) [right of=x] {$Y$};
    \node[vertex, circle, red, opacity=.7] (u) [above of=x, yshift=-0.2cm] {$U$};

    \draw[edge] (z) to (a);
    \draw[edge] (a) to (x);
    \draw[edge] (x) to (y);
    \draw[edge, red, opacity=.7] (u) to (a);
    \draw[edge, red, opacity=.7] (u) to (x);
    \draw[edge, red, opacity=.7] (u) to (y);

    \node[below of=a, yshift=0.3cm, xshift=.5cm] (l) {$(c)$};
  \end{scope}
  
  \begin{scope}[yshift=-2.3cm, xshift=4.0cm]
    \tikzstyle{vertex} = [
      draw, thick, ellipse, minimum size=4.0mm, inner sep=1pt
    ]

    \tikzstyle{edge} = [
      ->, blue, very thick
    ]

    \node[vertex, rectangle] (z2) {$z$};
    \node[vertex, circle] (z) [above of=z2, yshift=-0.2cm] {$Z$};
    \node[vertex, circle] (a) [right of=z] {$A^z$};
    \node[vertex, rectangle] (a2) [right of=z2] {$a$};
    \node[vertex, circle, gray] (x) [right of=a2] {$X^a$};
    \node[vertex, circle] (y) [right of=x] {$Y^a$};
    \node[vertex, circle, red, opacity=.7] (u) [above of=x, yshift=-0.2cm] {$U$};

    \draw[edge] (z2) to (a);
    \draw[edge] (a2) to (x);
    \draw[edge] (x) to (y);
    \draw[edge, red, opacity=.7] (u) to (a);
    \draw[edge, red, opacity=.7] (u) to (x);
    \draw[edge, red, opacity=.7] (u) to (y);

    \node[below of=a2, yshift=0.3cm, xshift=0.5cm] (l) {$(d)$};
  \end{scope}
\end{tikzpicture}
  }
  \caption{ (a) - (c) show example Bayesian networks. Black variables represent
    observed variables, grey variables represent unobserved variables we wish to
    make inferences on, and red variables represent unobserved variables with
    unknown cardinality. (d) shows a single world intervention graph (SWIG)
    corresponding to graph (c) under the intervention $Z=z$ and $A=a$. }
  \label{fig:iv}
\end{figure}
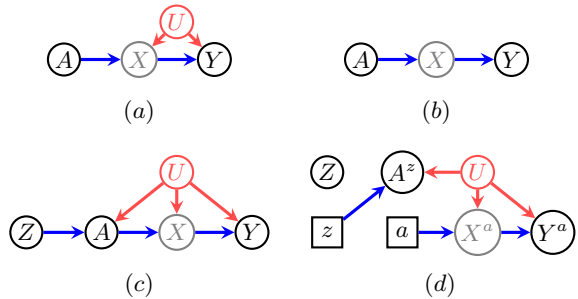

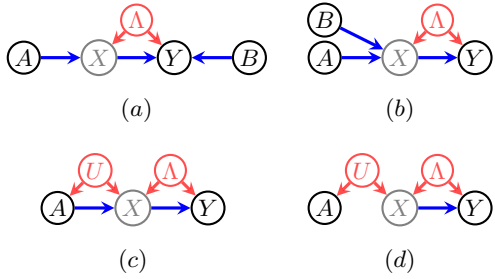
\begin{figure}[t]
  \centering {\small
    \begin{tikzpicture}[>=stealth, node distance=1.0cm]
  \begin{scope}[xshift=0.0cm]
    \tikzstyle{vertex} = [
      draw, thick, ellipse, minimum size=4.0mm, inner sep=1pt
    ]

    \tikzstyle{edge} = [
      ->, blue, very thick
    ]

    \node[vertex, circle] (a) {$A$};
    \node[vertex, circle, gray] (x) [right of=a] {$X$};
    \node[vertex, circle] (y) [right of=x] {$Y$};
    \node[vertex, circle, red, opacity=.7] (u) [above of=x, xshift=0.5cm, yshift=-0.5cm] {$\Lambda$};
    \node[vertex, circle] (b) [right of=y] {$B$};

    \draw[edge] (a) to (x);
    \draw[edge] (x) to (y);
    \draw[edge, red, opacity=.7] (u) to (x);
    \draw[edge, red, opacity=.7] (u) to (y);
    \draw[edge] (b) to (y);

    \node[below of=x, yshift=0.3cm, xshift=0.5cm] (l) {$(a)$};
  \end{scope} 
  
  \begin{scope}[xshift=4.0cm]
    \tikzstyle{vertex} = [
      draw, thick, ellipse, minimum size=4.0mm, inner sep=1pt
    ]

    \tikzstyle{edge} = [
      ->, blue, very thick
    ]

    \node[vertex, circle] (a) {$A$};
    \node[vertex, circle, gray] (x) [right of=a] {$X$};
    \node[vertex, circle] (y) [right of=x] {$Y$};
    \node[vertex, circle, red, opacity=.7] (u) [above of=x, xshift=0.5cm, yshift=-0.5cm] {$\Lambda$};
    \node[vertex, circle] (b) [above of=a, yshift=-0.5cm] {$B$};

    \draw[edge] (a) to (x);
    \draw[edge] (x) to (y);
    \draw[edge, red, opacity=.7] (u) to (x);
    \draw[edge, red, opacity=.7] (u) to (y);
    \draw[edge] (b) to (x);

    \node[below of=x, yshift=0.3cm] (l) {$(b)$};
  \end{scope}
  
  \begin{scope}[xshift=0.45cm, yshift=-2.0cm]
    \tikzstyle{vertex} = [
      draw, thick, ellipse, minimum size=4.0mm, inner sep=1pt
    ]

    \tikzstyle{edge} = [
      ->, blue, very thick
    ]

    \node[vertex, circle] (a) {$A$};
    \node[vertex, circle, red, opacity=.7] (u1) [above of=a, xshift=0.5cm, yshift=-0.5cm] {$U$};
    \node[vertex, circle, gray] (x) [right of=a] {$X$};
    \node[vertex, circle] (y) [right of=x] {$Y$};
    \node[vertex, circle, red, opacity=.7] (u2) [above of=x, xshift=0.5cm, yshift=-0.5cm] {$\Lambda$};

    \draw[edge] (a) to (x);
    \draw[edge] (x) to (y);
    \draw[edge, red, opacity=.7] (u2) to (x);
    \draw[edge, red, opacity=.7] (u2) to (y);
    \draw[edge, red, opacity=.7] (u1) to (a);
    \draw[edge, red, opacity=.7] (u1) to (x);

    \node[below of=x, yshift=0.3cm] (l) {$(c)$};
  \end{scope}
  
  \begin{scope}[yshift=-2.0cm, xshift=4.0cm]
    \tikzstyle{vertex} = [
      draw, thick, ellipse, minimum size=4.0mm, inner sep=1pt
    ]

    \tikzstyle{edge} = [
      ->, blue, very thick
    ]

    \node[vertex, circle] (a) {$A$};
    \node[vertex, circle, red, opacity=.7] (u1) [above of=a, xshift=0.5cm, yshift=-0.5cm] {$U$};
    \node[vertex, circle, gray] (x) [right of=a] {$X$};
    \node[vertex, circle] (y) [right of=x] {$Y$};
    \node[vertex, circle, red, opacity=.7] (u2) [above of=x, xshift=0.5cm, yshift=-0.5cm] {$\Lambda$};

    \draw[edge] (x) to (y);
    \draw[edge, red, opacity=.7] (u2) to (x);
    \draw[edge, red, opacity=.7] (u2) to (y);
    \draw[edge, red, opacity=.7] (u1) to (a);
    \draw[edge, red, opacity=.7] (u1) to (x);

    \node[below of=x, yshift=0.3cm] (l) {$(d)$};
  \end{scope}
\end{tikzpicture}
  }
  \caption{ (a) and (b) are classical instrumental graphs as described in
    Proposition \ref{prop:general_iv}, (c) and (d) are graphs whose constraints
    on the observed data distribution can also be represented linearly by
    Proposition \ref{prop:confounded_instrument}. }
  \label{fig:gen_iv}
\end{figure}

In the basic IV model (Figure \ref{fig:iv} (a)), the instrument $A$ is assumed
to be unconfounded with $X$; however, unconfoundedness is a strong
assumption that does not hold in a variety settings. Instead, we may be willing
to make a relaxed assumption that the confounders for $A$ and $X$ are
independent of the confounders for $X$ and $Y$, as shown in Figure
\ref{fig:gen_iv} (c). 
%
%
We can extend the class of graphs defined in Proposition \ref{prop:general_iv}
to include confounded instruments using the following special case of
Proposition 5 in \cite{evans}. An alternative proof of this proposition is
presented in the supplement.
\begin{proposition}
\label{prop:confounded_instrument}
  Suppose a vertex $A$ in a Bayesian network $\mathcal{G}$ has no observed or
  unobserved parents and has a single child $B$. Then the model for the
  variables in $\mathcal{G}$ is unchanged if an unobserved common parent of $A$
  and $B$ is added to the graph, or if the unobserved common parent is added and
  the edge from $A$ to $B$ is removed.
\end{proposition}
This proposition has two important consequences: first, instruments in
$\mathbf{A}$ may be confounded with their children and, second, if an
instrument is confounded with its child, it need not have a directed edge to
that child. This result broadens the set of graphical models for which the
constraints on the observed data distribution can be expressed linearly in
$\psi$ to include the graphs such as those shown in Figures \ref{fig:gen_iv}
(c) and (d). In particular, Figure \ref{fig:gen_iv} (d) can be used to represent a model where $A$ is a proxy for the true unobserved instrument.

Once we have expressed the modeling constraints as linear constraints relating
the parameters $\psi$ to the observed data conditional distributions $P({\bf x}
\mid {\bf a})$, we can now proceed exactly as in Section \ref{sec:base},
optimizing with respect to $\psi$ rather than $\phi$. Since $\phi$ is linearly
related to $\psi$, the observed data constraint and all measurement error
constraints from the previous section are still in linear in $\psi$ and can be
composed with the graphical constraints in this section. Alternatively, the
measurement error constraints can be expressed for each potential outcome,
representing a belief that these constraints hold in the observed data as well
as under various interventions.

\subsection*{Application: Bounding $E[X]$ in the IV model}

To demonstrate the use of graphical constraints, we will extend our example
from Section \ref{sec:base} to include an additional binary instrument $A \in
\{0,1\}$ and we will assume the graphical model in Figure \ref{fig:iv} (a). For
a complete description of the resulting LP, see Appendix \ref{app:iv_lp}.
Assume also that the observed conditional distributions $P(Y \mid A=a)$ for $a
= 0,1$ are the two marginal distributions shown in Figure \ref{fig:mrs_bounds}
and that $P(A=0) = \frac{1}{2}$. Then, using \emph{only} the constraints
encoded in the graph, we get the numerical bounds $E[X] \in [0.31,4.69]$. These
bounds can be made substantially tighter by including additional measurement
error constraints, but it is worth noting that we can achieve non-trivial
bounds by relying only on graphical constraints. \hfill\qedsymbol

\subsection{Computing bounds for non-linear models}

Unfortunately, many relevant models do not fall into the model class described
above. In this section, we describe how non-sharp outer bounds can be derived
for such cases. The only complete procedure for identifying all constraints
implied by Bayesian networks on the distribution of a subset of their vertices
is an application of quantifier elimination \cite{gieger}, which is infeasibly
slow for many problems. When constraints are known to exist, for example by
Evans' e-separation criterion \cite{evans2012eseparation}, their exact form may
not be known and may not be linear. When constraints are known, but are not
linear, it may be possible to derive sharp bounds analytically. For example,
the following proposition, proven in the supplementary material, gives sharp
bounds for a three variable Markov chain (Figure \ref{fig:iv} (b)) over binary
variables.
\begin{proposition}
    \label{prop:binary_bounds}
    Let $X$ and $Y$ be binary variables such that $X \not\independent Y$, let
    $A$ be a discrete variable such that $A \independent Y | X$ and $P(A=a) >
    0$ for all $a$, and let $p_y = P(Y=y)$ and $p_{y \mid a} = P(Y=y\mid A=a)$.
    Then we have the following sharp bounds on $P(X=1)$:
    \begin{align}
	  P(X=1) &\in \bigcup_{y\in\{0,1\}} \left[\frac{p_y - \min_a p_{y \mid a}}{1 - \min_a p_{y \mid a}},\frac{p_y}{\max_a p_{y\mid a}}\right]
    \end{align}
\end{proposition}
Such analytical bounds, however, are not typically available. In these cases,
\emph{non-sharp} bounds can be derived for any graph by first repeatedly
appealing to Proposition \ref{prop:confounded_instrument} and then adding
adding a latent confounder that meets the criteria of Proposition
\ref{prop:general_iv}. Specifically, any latent variable Bayesian network
$\mathcal{G}$ can be converted to a new graph $\mathcal{G}'$ that meets the
conditions of Proposition \ref{prop:general_iv} through the following steps:
\begin{enumerate}
\item For any latent confounder $U$ with exactly two children $A$ and $B$ such
  that $Pa(A) = \{U\}$ and $Ch(A) = \{B\}$ or $Ch(A) = \emptyset$, add an edge
  from $A$ to $B$ if it does not exist and remove $U$ from the graph.
\item Add a latent confounder $\Lambda$, and an edge from $\Lambda$ to each
  variable $V$ in the graph for which $Pa(V) \neq \emptyset$.
\end{enumerate}

\begin{figure}[t]
  \centering {\small
    \include{graph-relaxation}
  }
  \caption{ A non-linear graphical model $\mathcal G$ and an application of the
    linear relaxation procedure, resulting in the linear relaxed model $\mathcal
    G'$. }
  \label{fig:graph_relaxation}
\end{figure}

An example application of this procedure is shown in Figure
\ref{fig:graph_relaxation}. Because modifying the graph according to
Proposition \ref{prop:confounded_instrument} does not change the constraints on
$P({\bf X} | {\bf A})$, Step 1 of this procedure does not change the constraint
set. Further, adding an additional latent confound can only \emph{remove}
independencies from the graph, thus Step 2 represents relaxations of the
constraints on $\psi$. As a result, applying the LP approach to the resulting
model will result in \emph{outer bounds} on the true partial identification set
whose tightness will depend on how many edges were added in steps 2 and 3. In
the following section, we extend our earlier discussion of potential outcomes
to consider partial identification bounds for causal parameters and constraints
relating multiple potential outcomes.

\section{Causal parameters and constraints}
\label{sec:causal-bounds}

In the previous section, we used potential outcomes to reason about the
distribution of a mismeasured variable $X$. Suppose instead that we observe a
treatment variable $T$ and are directly interested in the potential outcome
$X(t)$ under some intervention $T = t$. As before, we do not observe $X$, but
instead observe a proxy $Y$. 
This scenario is common in fields like economics and epidemiology, in which the
treatment is exactly known, but the outcome is measured through inexact tools
such as surveys. In this section, we will show how the constraints presented in
the previous sections can be applied to target parameters involving the
distribution of $X(t)$ and will introduce additional constraints that apply
specifically to causal inference settings. We demonstrate this approach in two
important settings: a clinical trial with measurement error on the outcome, and
an IV model with measurement error on the outcome.

\subsection{Causal target parameters}

In order to use the constraints from the previous sections to bound causal
parameters, we first need to show how $\psi$, defined in the previous section,
can be linearly mapped to $P(X(t))$, and then how $P(X(t))$ can be linearly
mapped to various causal parameters. Assume a latent variable Bayesian network
$\mathcal{G}$ that meets the conditions of Proposition \ref{prop:general_iv}
and the corresponding joint distribution $\psi$ over instruments $\mathbf{A}$
and potential outcomes $\mathbf{\tilde{X}}$ is defined as in Section
\ref{sec:iv}. Then, for an arbitrary treatment variable $T \in \mathbf{A} \cup
\mathbf{X}$ and value $t$, we can calculate the distribution over $X(t)$ as in
a structural equation model, by intervening on $T$ in the graph and repeatedly
appealing to the consistency assumption to marginalize out all variables other
than $X(t)$ \cite{pearl2009causality}. For example, assuming the IV model in
Figure \ref{fig:iv} (a), the distribution $P(X(a))$ is just a marginal of
$\psi$ and the distribution $P(Y(a))$ can be derived as
\begin{align}
	P(Y(a) = y) = \sum_x P(Y(x) = y, X(a) = x) 
\end{align}
In general, for endogenous variable $X$ and intervention $T=T$, this expression
can be constructed as follows. Let $\bf \tilde V$ represent all potential
outcomes for variables other than $X$. The general expression is then
\begin{equation*}
P(X(t) = x) = \sum_{{\bf v}} P(X(g({\bf v}, T=t)) = x, {\bf V = v}),
\end{equation*}
%
where $g({\bf v}, T =t)$ is computed recursively as
\begin{align*}
  g&({\bf v}, T=t) = \\
  &\begin{cases}
    \mathbf{v}_{Pa(X)\setminus\{T\}}, T=t &\text{if~} X \in Ch(T)\\
    \{g({\bf v}, C(t)={\bf v}_{c(t)}) \text{~for~} C \in Ch(T)\} &\text{otherwise}.
  \end{cases}
\end{align*} 
This marginalization is linear in $\psi$, and thus any causal parameter that
can be written as a linear function of $P(X(t))$ can also be written as a
linear function of $\psi$. This includes the average treatment effect (ATE)
which is defined as $E[X(t) - X(t')]$ as well as the probability of non-zero
treatment effect, which can be written as $P(X(t) \neq X(t'))$. As in the
previous sections, we can express observed data, measurement error, and
graphical constraints covered by Proposition \ref{prop:general_iv} as linear
constraints on $\psi$, allowing us to compute bounds on target parameters
involving $P(X(t))$ under these constraints.

Importantly, the above mapping applies to the full class of graphs defined by
Proposition \ref{prop:general_iv}, but applying Proposition
\ref{prop:confounded_instrument} to this setting requires a bit more care. If
$T$ is in the set of instruments $\mathbf{A}$, then augmenting the graph as
described in Proposition \ref{prop:confounded_instrument} does, in fact, change
the constraints on $P(X(t))$. If, however, $T$ is in $\mathbf{X}$, then
augmenting the graph according to Proposition \ref{prop:confounded_instrument}
leaves the constraints on $P(X(t))$ unchanged, as all instruments still satisfy
the conditions of Proposition \ref{prop:confounded_instrument}.

\subsection{Causal Assumptions}

Finally, we may want to make additional \textbf{causal assumptions}, which
relate potential outcomes under different interventions. For example, assuming
we are interested in $X(t)$ and $X$ is proxied by $Y$, below are two commonly
made monotonicity assumptions which may be encoded as linear constraints.
\begin{enumerate}
  \item[(A4)] Positive Effect of Treatment on Truth:
    $P(X(t) = x, X(t') = x') = 0\,\, \forall t' > t, x' < x$
  \item[(A5)] Positive Effect of Truth on Proxy:
    $P(Y(x) = y, Y(x') = y') = 0\,\, \forall x' > x, y' < y$
\end{enumerate}
Assumption (A4) is appropriate if there are strong reasons to believe the
outcome under treatment $t$ will be strictly higher than $t'$.
Assumption (A5) is employed whenever it is assumed that, even under measurement
error, intervening to increase $X$ will lead to an increase in $Y$. Additional
causal constraints, such as limits on the effect size or the proportion
affected, may be similarly imposed. As with the measurement error
assumptions, these equality constraints can be relaxed by specifying that the
sums are bounded from above, rather than identically equal to zero.

\begin{figure}[t]
	\begin{subfigure}[b]{0.49\textwidth}
			\includegraphics[width=\textwidth]{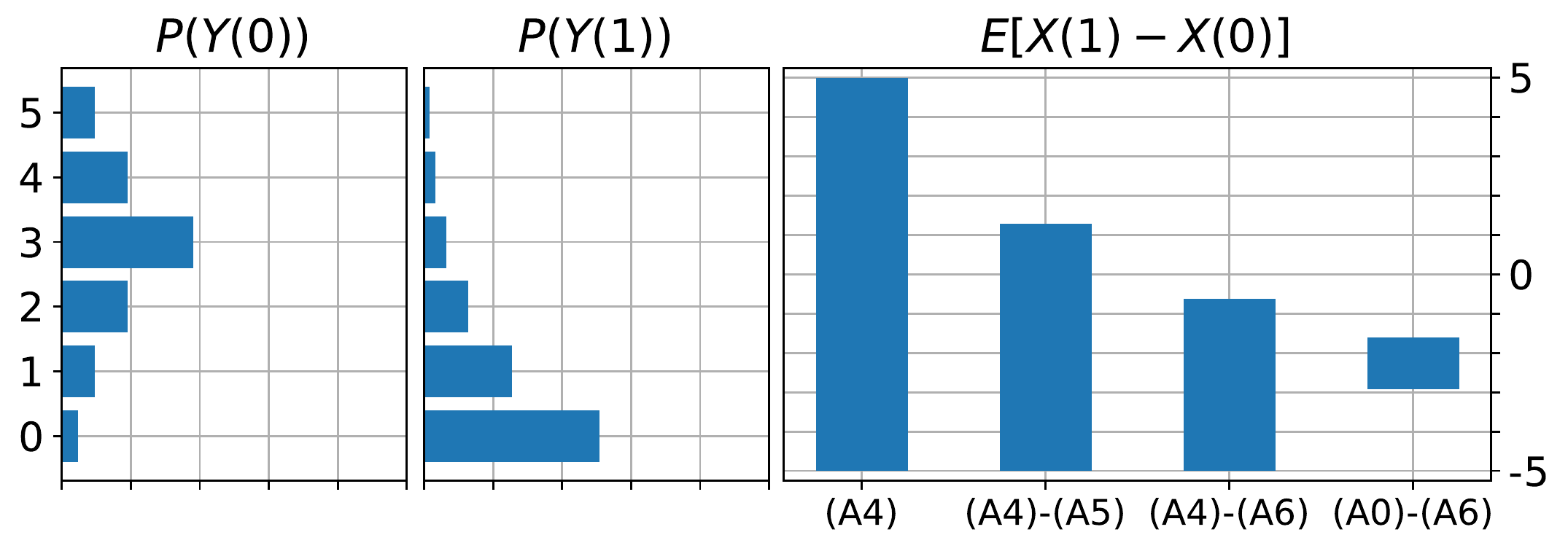}
			\caption{}
    \end{subfigure}
    \begin{subfigure}[b]{0.49\textwidth}
			\includegraphics[width=\textwidth]{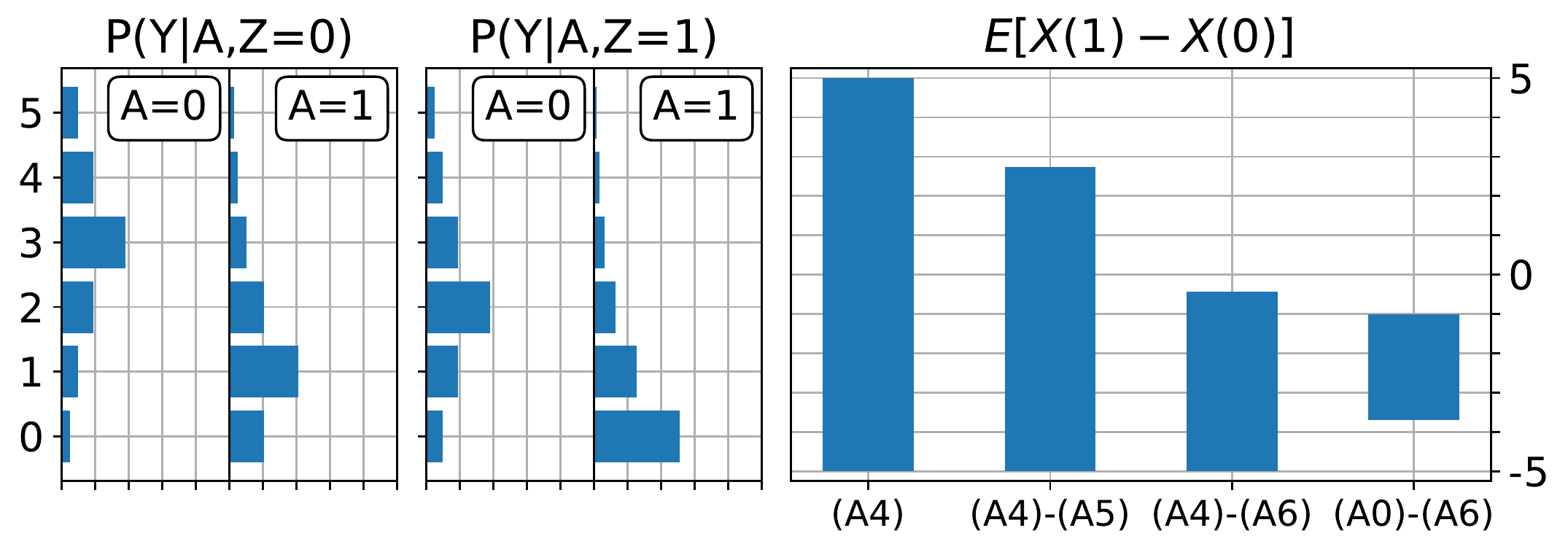}
			\caption{}
	\end{subfigure}\hspace{1cm}
  \caption{
    Bounds on the ATE $E[X(1) - X(0)]$ under different measurement
    error assumptions. (a) observed data distributions and bounds under in a
    randomized trial setting and (b) shows observed data distributions and
    bounds in an instrumental variable setting. Note that in the binary
    instrumental variable setting, $Z$ can be thought of as the assigned
    treatment and $A$ can be thought of as the observed treatment under
    partial compliance. $A=Z$ then corresponds to observed compliance and, in
    these cases ($A=Z=0$ and $A=Z=1$), the proxy distributions match the
    distributions in the randomized trial.
  } 
  \label{fig:causal_bounds}
\end{figure}

\subsection*{Application: Bounding the ATE in a randomized trial}

To demonstrate computation of bounds for a causal parameter, we will now extend
our previous examples to compute bounds on the ATE in a randomized trial where
there is measurement error on the outcome variable. Assume that we are
interested in the ATE of a binary treatment $A \in \{0,1\}$ on a variable $X$
using data from a single noisy proxy $Y$, with $X$ and $Y$ defined as before.
We will assume the graphical model shown in Figure \ref{fig:iv} (a) where $U$
represents an unobserved confounder. This model trivially satisfies the
conditions of Proposition \ref{prop:general_iv} and thus all graphical
constraints can be expressed linearly. Figure \ref{fig:causal_bounds} (a) shows
the resulting bounds on the ATE as additional constraints are added. With only
the graphical constraints, the bounds computed are the trivial bounds, though
this will not be the case for all target parameters. Adding the causal
assumptions (A4)-(A5), however, we are able to meaningfully bound the ATE away
from zero. This bound becomes much tighter when the measurement error
constraints are added. \hfill\qedsymbol

\subsection*{Application: Bounding the ATE in an instrumental variable model}

Assume now that we are interested in the ATE of $A$ on $X$, but that $A$ is no
longer randomized. Instead assume that we observe a binary instrumental
variable $Z \in \{0,1\}$ and that all variables follow the graphical model
shown in Figure \ref{fig:iv} (c) where again $U$ represents an unobserved
confounder. This model, which represents the IV model with measurement error on
the outcome variable, similarly satisfies the conditions of Proposition
\ref{prop:general_iv} and we can again express all graphical constraints
linearly. We assume that $P(A=0 \mid Z=0) = P(A=1 \mid Z=1) = 0.8$ and $P(Y
\mid A,Z)$ is distributed as shown in Figure \ref{fig:causal_bounds} (b). These
distributions were chosen to roughly simulate a trial with partial compliance
where $Z$ represents treatment assignment and $A$ represents actual treatment.
The resulting bounds on the ATE under different assumptions are also shown in
Figure \ref{fig:causal_bounds} (b). As expected, these bounds are wider than
those produced in the randomize trial case, but the ATE can nevertheless be
meaningfully bounded away from zero with only the causal assumptions (A4)-(A5).
\hfill\qedsymbol

We re-emphasize that no known symbolic bounds exist for either of these
settings and that the various numerical bounds in each setting were computed by
simply making small changes to the implied LP. For a full description of the
LPs in both of these cases, see Appendix \ref{app:causal_lp}.
\section{Related work} 

Measurement error occurs in many scientific settings and there is substantial
literature on identification spread across a number of different methodological
sub-disciplines. Much of this work concerns point identification in parametric
models and we refer the interested reader to \cite{carroll2006measurement} and
\cite{gustafson2003measurement} for full treatments of these topics. In this
section, we review several results on non-parametric partial identification in
measurement error and related settings.

Several works, particularly in econometrics, have presented partial
identifiability results under various measurement error models.
\cite{horowitz1995contaminated} consider the setting presented in Section
\ref{sec:base}, deriving sharp bounds on the distribution of the ground truth
under a particular error model where data is "contaminated" by data from
another, unknown, distribution. \cite{molinari2008misclassified} consider the
same setting, presenting a procedure for verifying whether a particular
distribution is in the identified set under a wide range of assumptions about
the error distribution, including some non-linear assumptions.
\cite{henry2014partial} consider partial identifiability in a class of finite
mixture models which includes, as a special case, the Markov chain model
considered in Proposition \ref{prop:binary_bounds}, similarly proposing a
method for verifying if a distribution is in the identified set. Our work
differs from \cite{molinari2008misclassified} and \cite{henry2014partial} in
that our methods do not require guess-and-check to calculate the complete
identified set.

 
The optimization-based approach we use to derive sharp bounds is inspired by
the approach used by \cite{balke93ivbounds} to derive sharp bounds on causal
effects in trials with partial compliance. This approach was similarly applied
by \cite{imai2010treatment} to partially identify the ATE under
measurement error on the treatment variable. This work is also related to
efforts to enumerate constraints on margins of latent variable Bayesian Networks
implied by the model \cite{wolfe2016inflation, evans2012eseparation,
  evans2018margins}. In such works, unobserved variables are not of primary
interest and do not have known cardinality, so no attempt is made to bound
functionals of their distribution. However, as indicated by our use of results
from \cite{bonet2001instrumentality}, constraints on the observed data law can
be used to derive restrictions on unobserved variables of known cardinality.


%
	
\section{Discussion} 

In this work, we presented an approach for computing bounds on distributional
and causal parameters involving a discrete variable which is subject to
measurement error. At the heart of this approach is the encoding of the target
parameter and modeling constraints as linear functions of the joint
distribution of all variables in the model. The target parameter can then be
maximized and minimized, with respect to this distribution and subject to the
the modeling constraints, to produce sharp bounds for any observed data
distribution. In particular, we provided a characterization of a class of
graphical models that can be linearly expressed, and a procedure for finding a
linear relaxation of models outside this class. We
applied our approach to produce bounds under measurement error in settings with
one or more proxies, including multiple important settings for which no known
bounds currently exist.

\bibliographystyle{plain}
\bibliography{references}

\appendix
\onecolumn
%
	
\section{Proof of Proposition \ref{prop:binary_bounds}}
\label{app:prop_1_proof}

\begin{proof}

	By assumption, $P(Y=1|X=0) \neq P(Y=0|X=0)$ and therefore

	\begin{align}
		\label{eq:px}
		P(X=1) = \frac{P(Y=1) - P(Y=1|X=0)}{P(Y=1|X=1) - P(Y=1|X=0)}
	\end{align}

    We will derive sharp bounds on $P(X=1)$ by taking the union of the sharp
    bounds when $P(Y=1|X=0) > P(Y=0|X=0)$ and when $P(Y=1|X=0) < P(Y=0|X=0)$.
    The RHS of Equation \ref{eq:px} is continuous on each of these sub-regions,
    thus, as in the the linear programming case, we can find sharp bounds on
    each sub-region by finding the maximum and minimum of $P(X=1)$ on each
    sub-region subject to the modeling constraints. Consider first the case
    when $P(Y=1|X=0) > P(Y=0|X=0)$. For each value $a \in \mathcal{A}$ we have
    the following constraint which combines the observed data constraint and
    the conditional independence assumption $A \independent Y | X$:
    
	\begin{align*}
	    P(Y=1|A=a) = &P(Y=1|X=0)(1-P(X=1|A=a))\\ &+ P(Y=1|X=0)P(X=1|A=a)
	\end{align*}
    
     Let $q_{y|x} := P(Y=y|X=x)$ and $\pi_{x|a} := P(X=x|A=a)$. Then, using
     Equation \ref{eq:px}, we can find the sharp upper bound for $P(X=1)$ on
     the $P(Y=1|X=0) > P(Y=0|X=0)$ sub-region by solving the following
     (non-linear) optimization problem:
    
    \begin{align*}
        \max_{q,\pi} \quad & \frac{p_1 - q_{1|0}}{q_{1|1} - q_{1|0}}\\
        \textrm{s.t.} \quad & p_{1|a} = q_{1|0}(1-\pi_{1|a}) + q_{1|1}\pi_{1|a}\,\,\,\forall a\\
		& q_{0|x} + q_{1|x} = 1\,\,\,\forall x\\
		& \pi_{0|a} + \pi_{1|a} = 1\,\,\,\forall a\\
        & 0 \leq q_{y|x},\pi_{x|a} \leq 1\,\,\,\forall y,x,a
    \end{align*}
    
    To solve this optimization problem, we will fix $q_{1|1}$ and optimize with
    respect to $q_{1|0}$ and then optimize the resulting function with respect
    to $q_{1|1}$. That is, let
	
    \begin{align*}
        g(q_{1|1}) = \max_{q_{1|0}} \quad & \frac{p_1 - q_{1|0}}{q_{1|1} - q_{1|0}}\\
        \textrm{s.t.} \quad & p_{1|a} = q_{1|0}(1-\pi_{1|a}) + q_{1|1}\pi_{1|a}\,\,\,\forall a\\
        & 0 \leq \pi_{1|a} \leq 1\,\,\,\forall a\\
        & 0 \leq q_{1|0} < q_{1|1}
    \end{align*}

    In this case, all constraints are satisfied if and only if $0 \leq q_{1|0}
    \leq \min_a p_{1|a}$ and the maximum is achieved when $q_{1|0} = 0$. Thus,
    $g(q_{1|1}) = \frac{p_1}{q_{1|1}}$. Next, we solve
    
    \begin{align*}
        \max_{q_{1|1}} \quad & g(q_{1|1}) = \frac{p_1}{q_{1|1}}\\
        \textrm{s.t.} \quad & p_{1|a} = q_{1|1}\pi_{1|a}\,\,\,\forall a\\
        & 0 \leq \pi_{1|a} \leq 1\,\,\,\forall a
    \end{align*}
    
    In this case, all constraints are satisfied if and only if $\max_a p_{1|a}
    \leq q_{1|1} \leq 1$ and the maximum value that satisfies this constraint
    is $\frac{p_1}{\max_a p_{1|a}}$. Applying similar reasoning to the
    minimization problem, we get a minimum value of $\frac{p_1 - \min_a
    p_{1|a}}{1 - \min_a p_{1|a}}$. Thus, when $q_{1|1} > q_{1|0}$, we have the
    following sharp bounds on $P(X=1)$
    
    \begin{align}
        \frac{p_1 - \min_a p_{1|a}}{1 - \min_a p_{1|a}} \leq P(X=1) \leq \frac{p_1}{\max_a p_{1|a}}
    \end{align}
    
    Finally, we repeat this derivation for $q_{1|1} < q_{1|0}$ and take the
    union of these two sets of bounds to get the bounds in Proposition
    \ref{prop:binary_bounds}. The bounds for the $q_{1|1} < q_{1|0}$ are simply
    one minus the bounds for $q_{1|1} < q_{1|0}$ and thus the bounds for
    $P(X=1)$ are the same as the bounds for $P(X=0)$\footnote{This is
    unsurprising as it reflects simple label-switching. In fact, the two
    sub-regions in this proof correspond to the two possible bipartite
    matchings between $X$ and $Y$ labels.}.

\end{proof}


%
	
\section{Combining Proposition \ref{prop:binary_bounds} with Measurement Error Assumptions}
\label{app:combinations_with_markov}

In the presence of additional measurement error assumptions, the bounds in Proposition \ref{prop:binary_bounds} can be further refined. In this section we present a few such refinements for measurement error assumptions (A1) and (A3) presented in Section \ref{sec:base} of the main paper as well as an additional non-linear assumption. (A2) does not apply to the binary case. All of these bounds can be derived with small modifications to the proof of Proposition \ref{prop:binary_bounds}.

\begin{corollary}
    If, in addition to the assumptions made in Proposition \ref{prop:binary_bounds}, we assume $P(Y=1|X=0) = 0$ (i.e. (A1)), we have the following sharp bounds
    
    \begin{align*}
        P(X=1) \in \left[p_1,\frac{p_1}{\max_a p_{1|a}}\right]
    \end{align*}
\end{corollary}

\begin{corollary}
    If, in addition to the assumptions made in Proposition \ref{prop:binary_bounds}, we assume $P(Y=0|X=0) > P(Y=1|X=0)$ (i.e. (A3)), we have the following sharp bounds
    
    \begin{align*}
        P(X=1) \in \left[\frac{p_1 - \min_a p_{1|a}}{1 - \min_a p_{1|a}},\frac{p_1}{\max_a p_{1|a}}\right]
    \end{align*}
\end{corollary}

\begin{corollary}
    If, in addition to the assumptions made in Proposition \ref{prop:binary_bounds}, we assume $P(Y=1|X=0) = P(Y=0 | X=1)$ (i.e. label independent noise), we have the following sharp bounds
    
    \begin{align*}
        P(X=1) \in \left[p_1,\frac{p_1 - p^*}{1 - 2p^*}\right] \cup \left[1-\frac{p_1 - p^*}{1 - 2p^*},1-p_1\right]
    \end{align*}
    
    where $p^* = \min \left\{\min_a p_{1|a},1-\max_a p_{1|a}\right\}$.
\end{corollary}


%
	
\section{Proof of Proposition \ref{prop:confounded_instrument}}
\label{app:prop_2_proof}
\begin{proof}
  We first address the case in which an unobserved common parent is added.
  The factorization of the distribution over observed variables implied by
  $\mathcal G$ and by the DAG that results after the addition of the common
  parent, denoted $\mathcal G'$ differ only in that the expression
  $P(A)P(B \mid pa_{\mathcal G}(B))$
  in the former is replaced with
  $\int_u P(U)P(A \mid U)P(B \mid pa_{\mathcal G}(B), U)$.
  We now show that these expressions are equivalent.

\begin{align*}
  \int_u &P(U)P(A \mid U)P(B \mid pa_{\mathcal G}(B), U)\\
  &= \int_u P(U \mid pa_{\mathcal G}(B))P(A \mid U, pa_{\mathcal G}(B) \setminus \{A\})P(B \mid pa_{\mathcal G}(B), U)\\
  &= \int_u P(U, A, B \mid pa_{\mathcal G}(B) \setminus \{A\})\\
  &= P(A, B \mid pa_{\mathcal G}(B) \setminus \{A\})\\
  &= P(A \mid pa_{\mathcal G}(B) \setminus \{A\}) P(B \mid pa_{\mathcal G}(B))\\
  &= P(A) P(B \mid pa_{\mathcal G}(B))
\end{align*}

The first step is due $\{A, U\} \perp_d pa_{\mathcal G}(B) \setminus \{A\}$ in
$\mathcal G'$, because by construction $B$ is a collider on all paths between
$\{A, U\}$ and $pa(B) \setminus \{A\}$. The second step is due to the chain rule
of probabilities. The next two steps are a simple marginalization of $U$ and an
expansion according to the chain rule, respectively. The final step is due again
to $A \perp_d pa_{\mathcal G}(B) \setminus \{A\}$ in $\mathcal G'$. This shows
that after marginalization of the added common parent of $A$ and $B$, we recover
exactly the model Markov to $\mathcal G$ over the variables in $\mathcal G$.

We now consider the scenario in which we add a common parent of $A$ and $B$, and
remove the directed edge from $A$ to $B$. We denote the resulting DAG by
$\mathcal G''$. The proof procedes very similarly; we show that
$P(A)P(B \mid pa_{\mathcal G}(B))$
is equivalent to
$\int_u P(U)P(A \mid U)P(B \mid pa_{\mathcal G}(B) \setminus \{A\}, U)$. To do
so, we note that $B \perp_d A \mid U, pa_{\mathcal G}(B)$ in $\mathcal G''$, as
$A$ has no parents other than $U$, yielding

\begin{align*}
  \int_u &P(U)P(A \mid U)P(B \mid pa_{\mathcal G}(B) \setminus \{A\}, U)\\
  &= \int_u P(U)P(A \mid U)P(B \mid pa_{\mathcal G}(B), U).
\end{align*}

We can now procede exactly as before, concluding the proof.
\end{proof}

%
	
\section{Linear program for bounding $E[X]$ with two proxies}
\label{app:iv_lp}

Below is the full linear program used for the application example in Section
\ref{sec:iv} of the main paper. The distribution $\psi$ is over the variables
$\tilde{X} = \{X(a): a \in \mathcal{A}\}$ and $\tilde{Y} = \{Y(x) : x \in
\mathcal{X}\}$. This LP includes only the probability constraints and the
observed data constraint and, as described in Section \ref{sec:iv}, the
graphical constraints are implicitly enforced by $\psi$.

\begin{align}
\label{eq:iv_lp}
&\text{objective:~~} &\sum_{\tilde{x},\tilde{y}} (P(A=0) \tilde{x}_0 &\psi_{\tilde{x}\tilde{y}} 
	+  P(A=1) \tilde{x}_1 \psi_{\tilde{x}\tilde{y}})\\\nonumber
&\text{constraints:~~} &\sum_{\tilde{x},\tilde{y}} [\tilde{y}_{\tilde{x}_0} = y] \psi_{\tilde{x}\tilde{y}} & = P(Y = y \mid A = 0) \\\nonumber
& &\sum_{\tilde{x},\tilde{y}} [\tilde{y}_{\tilde{x}_1} = y] \psi_{\tilde{x}\tilde{y}} & = P(Y = y \mid A = 1) \\\nonumber
& &\psi_{\tilde{x}\tilde{y}} & \ge 0 \\\nonumber
\end{align}


%

\section{Linear Program for Causal Bounds}
\label{app:causal_lp}

First, we present the linear program used to bound the ATE in a randomized
trial. As in the previous example, the distribution $\psi$ is over the
variables $\tilde{X} = \{X(a): a \in \mathcal{A}\}$ and $\tilde{Y} = \{Y(x) : x
\in \mathcal{X}\}$. We include assumptions (A0), and (A2) through (A6), as
described in the example given in Section \ref{sec:causal-bounds} of the main
paper. To obtain an LP corresponding to a subset of these assumptions, this LP
can be modified by dropping constraints corresponding to assumptions not in
that subset. Note that each of the constraints on measurement error are
repeated twice - once to place restrictions on the relationship between $X(a)$
and $Y(a)$, and once for $X(a')$ and $Y(a')$. Then the ATE can be bounded by
solving the LP in Figure \ref{fig:ate_lp}.

\begin{figure}[t!]
\begin{align}
\label{eq:rct-ate}
  &\text{objective:~~} &\sum_{\tilde{x},\tilde{y}}(\tilde{x}_1 - \tilde{x}_0)\psi_{\tilde{x}\tilde{y}} \\\nonumber
  &\text{constraints:~~}
  &\psi_{\tilde{x}\tilde{y}} &\ge 0                                  &\forall\, \tilde{x},\tilde{y}~~~~(PROB) \\\nonumber
&  &\sum_{\tilde{x},\tilde{y}} [\tilde{y}_{\tilde{x}_a} = y] \psi_{\tilde{x}\tilde{y}} & = P(Y = y \mid A = a)   &\forall\, a,y~~~~(OBS)\\\nonumber
&  &\sum_{\tilde{x},\tilde{y}} [|\tilde{y}_{\tilde{x}_a} - \tilde{x}_a| > 2]\psi_{\tilde{x}\tilde{y}} &\le \epsilon &\forall\, a~~~~(A0)\\\nonumber
&  &\left|\sum_{\tilde{x},\tilde{y}} \psi_{\tilde{x}\tilde{y}}[\tilde{x}_a = x]([\tilde{y}_{\tilde{x}_a} = y] -  [\tilde{y}_{\tilde{x}_a} = y'])\right| &\le \lambda &\forall\, a, |x - y| = |x - y'| ~~~~(A2)\\\nonumber
&  &\sum_{\tilde{x},\tilde{y}} \psi_{\tilde{x}\tilde{y}} [\tilde{x}_a = x]([\tilde{y}_{\tilde{x}_a} = y] - [\tilde{y}_{\tilde{x}_a} = y'])&\ge 0 &\forall\, a, |x - y| > |x - y'| ~~~~(A3)\\\nonumber
&  &\sum_{\tilde{x},\tilde{y}} \psi_{\tilde{x}\tilde{y}} [\tilde{x}_1 < \tilde{x}_0] &= 0  &(A5)\\\nonumber
&  &\sum_{\tilde{x},\tilde{y}} \psi_{\tilde{x}\tilde{y}} [\tilde{y}_x < \tilde{y}_{x'}] &= 0  &\forall x > x'~~~~(A6)\\\nonumber
\end{align}
\caption{LP for bounding the ATE in a randomized trial.}
\label{fig:ate_lp}
\end{figure}

Next, we consider the LP used for bounding the ATE in the IV setting. We
describe this LP in relation to the LP in Figure \ref{fig:ate_lp}. In this
case, $\psi$ is now also a distribution over $\tilde{A} = \{A(z): z \in
\mathcal{Z}\}$. Then objective then becomes
$\sum_{\tilde{a},\tilde{x},\tilde{y}}(\tilde{x}_1 -
\tilde{x}_0)\psi_{\tilde{a}\tilde{x}\tilde{y}}$. The potential outcomes $A(z)$
and $A(z')$ are similarly marginalized out of assumptions (A0), and (A2)
through (A6). The only substantive change is that the observed data constraints
must be modified as described in Section \ref{sec:causal-bounds}. This
modification yields the following constraints

\begin{align*}
	\sum_{\tilde{a},\tilde{x},\tilde{y}} \psi_{\tilde{a}\tilde{x}\tilde{y}}[\tilde{a}_z = a,\tilde{y}_{\tilde{x}_a} = y] &= P(Y = y, A = a \mid Z = z) &\forall\, z, a, y.
\end{align*}


\end{document}